\documentclass[runningheads]{llncs}
\usepackage[utf8]{inputenc}
\usepackage[textsize=footnotesize]{todonotes}
\usepackage{epstopdf}
\usepackage{graphicx}
\usepackage{subcaption}
\usepackage{caption}
\usepackage{algpseudocode}
\usepackage{algorithm} 
\usepackage{enumitem}
\usepackage{booktabs}
\usepackage{amsmath, amssymb}
\usepackage{amssymb}
\usepackage{comment}
\usepackage{verbatimbox}
\usepackage{multirow}
\usepackage{threeparttable}

\DeclareMathOperator*{\argmin}{arg\,min}

\title{Boosted Embeddings for Time-Series Forecasting}
\author{
Sankeerth Rao Karingula \and Nandini Ramanan \and Rasool Tahmasbi \and Mehrnaz Amjadi \and
Deokwoo Jung \and Ricky Si \and  Charanraj Thimmisetty \and Luisa Polania Cabrera \and Marjorie Sayer \and Claudionor Nunes Coelho Jr
}
\institute{ADVANCED APPLIED AI RESEARCH\\
PALO ALTO NETWORKS\\
\url{https://www.paloaltonetworks.com/}}

\date{March 2021}

\begin{document}

\maketitle
\begin{abstract}
Time-series forecasting is a fundamental task emerging from diverse data-driven applications. Many advanced autoregressive methods such as ARIMA were used to develop forecasting models. 
Recently, deep learning based methods such as DeepAR, NeuralProphet, Seq2Seq have been explored for the time-series forecasting problem. In this paper, we propose a novel time-series forecast model, \texttt{DeepGB}.
We formulate and implement a variant of gradient boosting wherein the weak learners are deep neural networks whose weights are incrementally found in a greedy manner over iterations. In particular, we develop a new embedding architecture that improves the performance of many deep learning models on time-series data using a gradient boosting variant.
We demonstrate that our model outperforms existing comparable state-of-the-art methods using real-world sensor data and public data sets.

\end{abstract}

\section{Introduction}
Time-series forecasting plays a key role in many business decision-making scenarios and is one of the central problems in engineering disciplines. In particular, many prediction problems arising in financial data\cite{finance}, weather data\cite{weather}, econometrics\cite{econometrics} and medical data\cite{healthcare} can be modeled as a time-series forecasting problem. 

Time-series forecasting models can be developed using various autoregressive (AR) methods.
Classical linear models such as autoregressive integrated moving average (ARIMA)~\cite{bobjenkins} are used to explain the past behavior of a given time-series data and then used to make predictions of the time-series. ARIMA is one of the most widely used forecasting methods for univariate time-series data forecasting. To account for seasonality in time-series data, ARIMA models can be further extended to seasonal autoregressive integrated moving average (SARIMA)~\cite{bobjenkins}.
In turn, SARIMA models can be extended with covariates or other regression variables to Seasonal AutoRegressive Integrated Moving Averages with eXogenous regressors, referred to as SARIMAX model, where the X added to the end stands for “exogenous”. The exogenous variable can be a time-varying measurement like the inflation rate, or a categorical variable separating the different days of the week, or a Boolean representing special festive periods. One limitation of AR models is that they become impractically slow when attempting to model long-range dependencies and do not scale well for large volumes of training data due to the strong assumptions they impose on the time-series~\cite{triebe2019ar}.

Recurrent neural networks (RNNs), and in particular Long Short Term Memory (LSTM) networks, have achieved success in time-series forecasting due to their ability to capture long-range dependencies and to model nonlinear functions. Hewamalage et. al. ~\cite{hewamalage2021recurrent} ran an extensive empirical study of the existing RNN models for forecasting. They concluded that RNNs are capable of modeling seasonality directly if the series contains homogeneous seasonal patterns; otherwise, they recommended a deseasonalization step. They demonstrated that RNN models generally outperform ARIMA models. However, RNN models require more training data than ARIMA models as they make fewer assumptions about the structure of the time series and they lack interpretability.

A popular RNN-based architecture is the Sequence to Sequence (Seq2Seq) architecture~\cite{hwang2019novel}, which consists of the encoder and the decoder, where both act as two RNN networks on their own. The encoder uses the encoder state vectors as an initial state, which is how the decoder gets the information to generate the output. The decoder learns how to generate target $y[t+1, \ldots]$ by matching the given target $y[\ldots, t]$ to the input sequence. The DeepAR model for probabilistic forecasting, recently proposed by Salinas et. al.~\cite{salinas2020deepar}, uses a Seq2Seq architecture for prediction. DeepAR can learn seasonal behavior and dependencies on given covariates and makes probabilistic forecasts in the form of Monte Carlo samples with little or no history at all. Also, it does not assume Gaussian noise and the noise distribution can be selected by users.

In an effort to combine the best of traditional statistical models and neural networks, the AR-Net was proposed~\cite{triebe2019ar}. It is a network that is as interpretable as Classic-AR but also scales to long-range dependencies. It also eliminates the need to know the true order of the AR process since it automatically selects the important coefficients of the AR process.  In terms of computational complexity with respect to the order
of the AR process, it is only linear for AR-Net, as compared to quadratic for Classic-AR.

Facebook Prophet~\cite{taylor2018forecasting} is another forecasting method, which uses a decomposable time-series model with three main model components: trend, seasonality, and holidays. Using time as a regressor, Prophet attempts to fit several linear and non-linear functions of time as components. Prophet frames the forecasting problem as a curve-fitting exercise rather than explicitly looking at the time-based dependence of each observation within a time series. NeuralProphet~\cite{NeuralProphet}, which is inspired by Facebook Prophet~\cite{taylor2018forecasting} and AR-Net~\cite{triebe2019ar}, is a neural network-based time-series model. It uses Fourier analysis to identify the seasons in a particular time series and can model trends (autocorrelation modeling through AR-Net), seasonality (yearly, weekly, and daily), and special events.


Inspired by the success of learning using {\em Gradient Boosting} (GB)~\cite{Friedman2002StochasticGB} and {\em Deep Neural Networks}~\cite{DeepLearningBook2016,wen2016cat2vec}, we present a novel technique for time-series forecasting called \underline{Deep} Neural Networks with  \underline{G}radient \underline{B}oosting or \texttt{DeepGB}. 
Neural networks can represent non-linear regions very well by approximating them with a combination of linear segments. It is proven by the universal approximation theorem~\cite{universal-theorem} that neural networks can approximate any non-linear region but it could potentially need an exponential number of nodes or variables in order to do this approximation. On the other hand, methods such as gradient boosting that build decision trees are very good at representing non-linear regions but they cannot handle complex arithmetic relations, such as multiplication of elements. The main idea of the proposed approach is to combine the strengths of gradient boosting and neural networks.




The proposed approach consists of building a series of regressors or classifiers and solving the problem for the residual at each time. This approach enables the generation of a number of small parallel models for a single task, instead of creating a large deep learning model that attempts to learn very complex boundary regions. Because each subsequent model attempts to solve only the gradient of the loss function, the task becomes simpler than attempting to perform a full regression or classification on the output range. Eventually, the error becomes small enough and indistinguishable from the data noise. 

Our main contributions are as follows.
\begin{itemize}
  \item We propose \texttt{DeepGB}, an algorithm for learning temporal and non-linear patterns for time-series forecasting by efficiently combining neural network embedding and gradient boosting.
  \item  We propose \textit{boosted embedding}, a computationally efficient embedding method that learns residuals of time-series data by incrementally freezing embedding weights over categorical variables. 
  \item In our empirical evaluation, we demonstrate how the proposed approach \texttt{DeepGB} scales well when applied to standard domains and outperforms state-of-the-art methods including \texttt{Seq2Seq} and \texttt{SARIMA} in terms of both efficiency and effectiveness. 
\end{itemize}

The rest of the paper is organized as follows: First, the background on gradient boosting is introduced in Section 2. Section 3 outlines the need for neural network embedding for time-series modelling, followed by the detailed description of the proposed approach, \texttt{DeepGB}. In Section 4, we present the experimental evaluation of the proposed method. Finally, conclusions and directions for future work are presented in Section 5.

\section{Gradient boosting}

In Gradient Boosting, the solution is comprised of simpler parallel models that are trained sequentially but added together at the end. One of the main ideas of gradient boosting is that each subsequent model (which may be as small as a small tree or as large as a full deep neural network) needs to only solve for the residue between the output and the previous regressors built, thus making it a much easier problem to solve.
Gradient Boosting~\cite{hastie2009boosting} is motivated by the intuition that finding multiple weak hypotheses to estimate local probabilistic predictions can be easier than finding a highly accurate model. Friedman~\cite{hastie2009boosting} in the seminal work, proposed a gradient boosting framework to train decision trees in sequence, such that each tree is modeled by fitting to the gradient/error from the tree in the previous iteration. Consider for $Data$ = $\{\langle \mathbf{x}, y \rangle \}$, in the gradient boosting process, a series of approximate functions $F_m$ are learned to minimize the expected loss $Loss:= E_x[L(y, F_m(x))]$ in a greedy manner:
\begin{align*}
    F_{m} := F_{m-1} + \rho_m * \psi_{m}
\end{align*}
where $\rho$ is the step size and $\psi_{m}$ is a tree selected from a series of $\mathbf{\Psi}$ functions to minimize $L$ in the $m$-th iteration:
\begin{align*}
    \psi_{m} = \argmin_{\psi\in\mathbf{\Psi}}E[L(y, F^{m-1}(x))+\mathbf\psi(x)]
\end{align*}
where the loss function is usually least-squares in most works and a negative gradient step is used to solve for the minimization function. 

In recent years three highly efficient and more successful gradient-based ensemble methods became popular, namely, XGBoost, LightGBM, and  CatBoost~\cite{DBLP:conf/nips/KeMFWCMYL17,makridakis2020m5}. However, both XGBoost and LightGBM suffer from overfitting due to biased point-wise gradient estimates. That is, gradients at each iteration are estimated using the same instances that were used by the current model for learning, leading to a bias. Unbiased Boosting with Categorical Features, known as CatBoost, is a machine learning algorithm that uses gradient boosting on decision trees~\cite{Dorogush2018CatBoostGB}. CatBoost gains significant efficiency in parameter tuning due to the use of trees that are balanced to predict labels. The algorithm replaces the gradient step of the traditional gradient boosting with ordered boosting, leading to reduced bias in the gradient estimation step. It additionally transforms the categorical features as numerical characteristics by quantization, i.e., by computing statistics on random permutations of the data set and clustering the labels into new classes with lower cardinality. In this paper, we employ CatBoost, which is known to demonstrate effectiveness from literature, as the gradient boosting algorithm of choice. 



In the field of Neural Networks (NNs), He et al.(\cite{he2016deep}) introduced a deep Residual Network (ResNet) learning architecture where trained ResNet layers are fit to the residuals. Although ResNet and gradient boosting are methods designed for different spaces of problems, there has been significant research that has gone into formalizing the gradient boosting perspective of ResNet~\cite{nitanda2018functional,veit2016residual}. Inspired by the complementary success of deep models and gradient boosting, we in this work propose gradient boosting of NNs, \texttt{DeepGB}, where we fit a sequence of NNs using gradient boosting. NNs have previously been used as weak estimators although the methods mostly focused on majority voting for classification tasks~\cite{hansen1990neural} and uniform or weighted averaging for regression tasks~\cite{opitz1996actively,perrone1992networks}. 
However, we are the first to investigate gradient boosting of NNs in the context of time-series forecast in this work.

\begin{algorithm}[!ht]
\begin{algorithmic}[1]
\Function{GradientBoosting}{$Data$,$M$,$Models$}
\State where $Data$ = $\{\langle \mathbf{x}_i, y_i \rangle \}$, $M$: data samples
\State $models$ = []
\State $F_0$ := $y$
\For {$1 \leq m \leq |Models|$} \Comment{Iterate over the Models}
\State $model$ = $Models.select()$
\State $model.fit(x, F_{m-1})$ \Comment{Fit the selected model}
\State $F_{m}$ = $F_{m-1} - model.predict(x)$ \Comment{Compute the residual}
\If {$abs(F_{m} - F_{m-1}) < \epsilon$} \Comment{Check termination condition}
    \State \bf{break}
\EndIf
\State $models.append(model)$ 
\EndFor
\State \Return $models$
\EndFunction
\end{algorithmic}
\caption{Gradient Boosting Algorithm}
\label{algo:gbm}
\end{algorithm}

Algorithm~\ref{algo:gbm} describes a generic algorithm of gradient boosting. 
It shows how a combination of simple models (but not trees) can be composed to create a better model by just adding them together, effectively assuming each subsequent model can solve the difference. Interestingly enough, nonlinear autoregressive moving average model with exogenous inputs - NARMAX~\cite{narmax} has proposed this approach by clustering increasingly more complex models together.
Note that Algorithm~\ref{algo:gbm} does not make any distinction between how to select the order of the models. In NARMAX~\cite{narmax}, for example, it is suggested that we should seek simpler models first, and start using more complex models later on.
\section{DeepGB Algorithm} \label{sec:boosts_emb}
In this section, we present our main algorithm \texttt{DeepGB} that is a combination of gradient boosting with embeddings.

It is often not straightforward to select the best model $F_{1} \cdots F_{m}$ in Algorithm~\ref{algo:gbm} as the models need to account for complex temporal dependency and non-linear patterns in the time-series. 
Our approach is to use a neural network with an embedding layer to learn the underlying residual features from data with time categorical variables extracted from variable time stamps. The embedding layers are introduced to learn distinct time-series at once by embedding their IDs, or to encode categorical features in a meaningful way (e.g., time-related variables like a month, week of the year, day of the week, day of the month, holidays, etc.) in a lower dimension and extract valuable information.

The first modification we will perform in the basic Gradient Boosting algorithm using general models is to consider 
$$
Models = \bf{list}(Embedding_1, \ldots, Embedding_N, Residual),
$$ 
where $Embedding_i$ is an embedding model that models categorical data, such as dates, holidays, or user-defined categories, and $Residual$ is a traditional machine learning model, such as deep neural networks, Gradient Boosting or Support Vector Machines.

An example of an embedding model can be seen in Figure~\ref{fig:keras} using Keras. In this example,  $\tt{embedding\_size}$ can be computed experimentally as \newline
$(\tt{number\_of\_categories}+1)^{0.25}$~\cite{blog-google-2017} or as $\min(50, (\tt{number\_of\_categories} + 1)//2)$~\cite{blog-ref-jeremy-howard}, where $1$ was added to represent one more feature that is required by the Keras Embedding layer.

\begin{verbbox}
categorical_input_shape = (window_size, 1)
embedding_size = min(50, (number_of_categories+1)/ 2)
x_inp = Input(categorical_input_shape)
x_emb = Embedding(
    number_of_categories + 1,
    embedding_size,
    input_length=window_size)(x_inp[..., -1:, 0])
x_out = Dense(1)(x_emb)
\end{verbbox}
\begin{figure}[ht]
  \centering
  \theverbbox
  \caption{Keras model for single output dense layer}
  \label{fig:keras}
\end{figure}

A standard gradient boosting approach to train these simple embedding models, $Embedding_1, \ldots, Embedding_N$, is to select one of those models at each iteration, train it and then compute its residual error. In the next iteration, the next selected embedding model is trained on such residual error.

\begin{figure}[!ht]
	\centering
	\includegraphics[width=\textwidth]{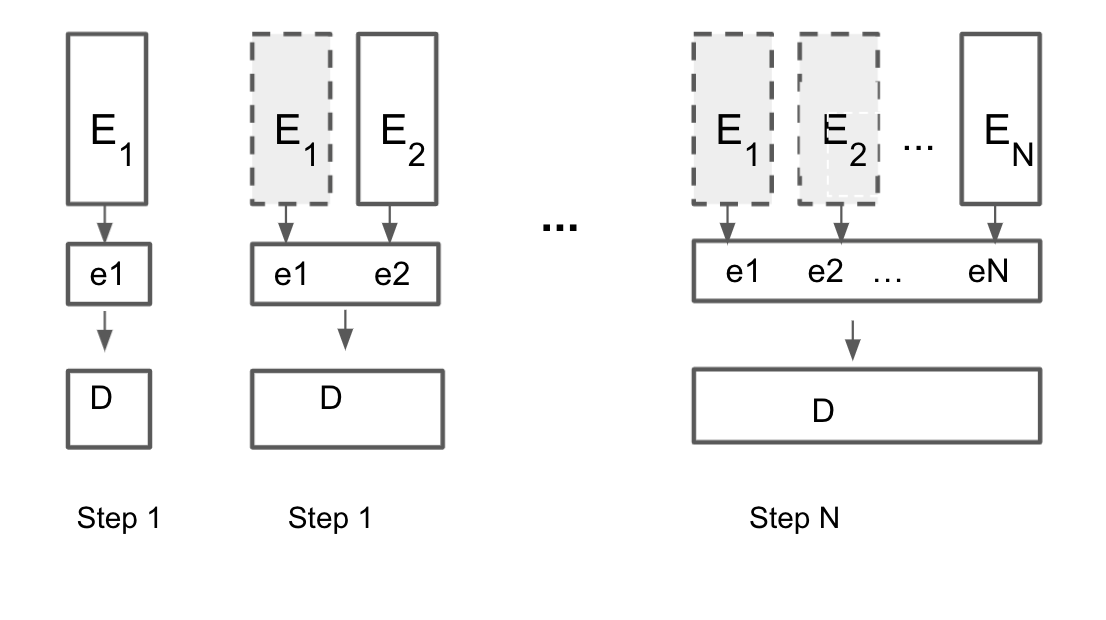}
	\caption{Training sequence by freezing of the embedding layers}
	\label{fig:emb}
\end{figure}

The key insight of this paper is that we can greatly reduce computational complexity by freezing embedding weights as we iterate through embedding training. The motivation behind this iterative freezing is that we preserve the gradient boosting notion of training weak learners on residuals. Each time an embedding is learned, we remove the effect of that embedding from the target data and train the next iteration on the residual. This idea can be best understood from Figure \ref{fig:emb}. 

In Figure \ref{fig:emb}, we are showing the sequential process in which the different embedding layers are trained. In particular, at any step $i$, we are essentially freezing the weights of the embedding layers $E_1,E_2,\ldots, E_{i-1}$ and then training the full model that concatenates all the $i$ embedding layers and applies the dense layers, D, on the concatenated output.

\paragraph{\texttt{DeepGB} Algorithm description:}
Pseudocode for the \texttt{DeepGB} algorithm is presented in Algorithm \ref{algo:bem}. The steps of the algorithm are described as follows
\begin{itemize}
    \item Line 6: We first initiate a sequence of embedding models, $E_1, \ldots, E_N$.
    \item Line 7: We specify the input shape of the categorical variable that goes into the current embedding model. 
    \item Line 8: We concatenate the current embedding with the previously concatenated embedding models.
    \item Line 10 - 13: We add dense layers that take the concatenated output of the embedding models as input. 
    \item Line 15: In the current iteration, the model learns the weights of the current embedding layer while the previous embedding models are frozen.
    \item Line 18: We don't need to learn for all N iteration and stop once the error converges below a certain threshold $\epsilon$.
    \item Line 21: The current frozen weights of the embedding model are appended to the embedding list. 
    \item Line 24: This is where the external model gets trained. In this paper, CatBoost is used as the external model.
\end{itemize}
 
\begin{algorithm}[h]
\begin{algorithmic}[1]
\Function{\texttt{DeepGB}}{$Data$,$M$,$Models = \left[E_1, \ldots, E_N, Residual\right]$}
\State where $Data$ = $\{\langle \mathbf{x}_i, y_i \rangle \}$, $M$: data samples
\State models = [], embeddings = []
\State $F_0$ = $y$
\For {$1 \leq m \leq N$} \Comment{Iterate over N embedding layers}
\State $c, embedding$ = $Models.select(E_1, \ldots, E_N)$ \Comment{Select an embedding layer at position $c$}
\State $\tt{x\_inp} = \tt{Input(categorical\_input\_shape)}$
\State $\tt{x\_emb} = \tt{Concatenate}(embeddings + \left[embedding(x\_inp[\ldots, -1:, c])\right])$ \\\Comment{Adding the next embedding layer}
\For {$1 \leq d \leq num\_dense\_layers$}
    \State $\tt{x\_emb} = \tt{Dense}(size, activation)(\tt{x\_emb})$\Comment{Adding dense layers}
\EndFor
\State $\tt{x\_out} = \tt{Dense}(1)(\tt{x\_emb})$
\State $model$ = $\tt{Model}(x\_inp, x\_out)$
\State $model.fit(x, y)$
\State Freeze weights of $embedding$ 
\State $F_{m}$ = $F_{m-1} - model.predict(x)$
\If {$abs(F_{m} - F_{m-1}) < \epsilon$}\Comment{Check termination condition}
\State {\bf break}
\EndIf
\State $embeddings.append(embedding)$
\EndFor
\State $models$ = $\left[model\right]$
\State $Residual.fit(x, F_{m})$ \Comment{Fit Residual (Catboost) model on the error}
\State $models.append(Residual)$
\State \Return $models$
\EndFunction
\end{algorithmic}
\caption{\texttt{DeepGB} Algorithm}
\label{algo:bem}
\end{algorithm}

\begin{figure}[!ht]
	\centering
	\begin{subfigure}[b]{1.0\textwidth}
		\includegraphics[width=\textwidth]{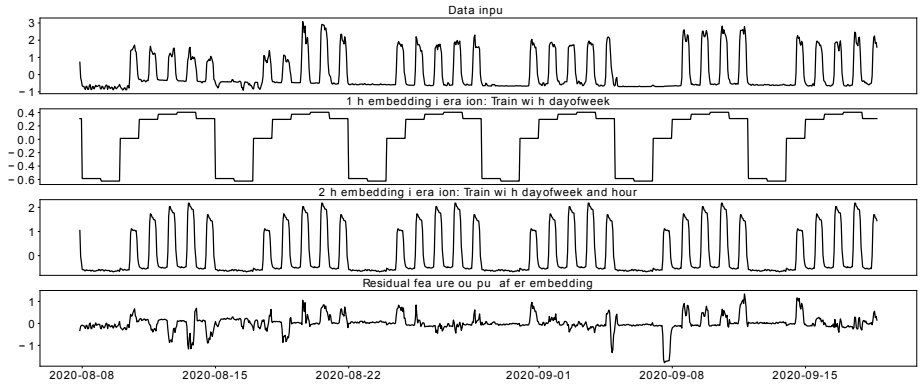}
		\caption{Data set 1}
		\label{fig:roc_1_a}
	\end{subfigure}
	\begin{subfigure}[b]{1.0\textwidth}
		\includegraphics[width=\textwidth]{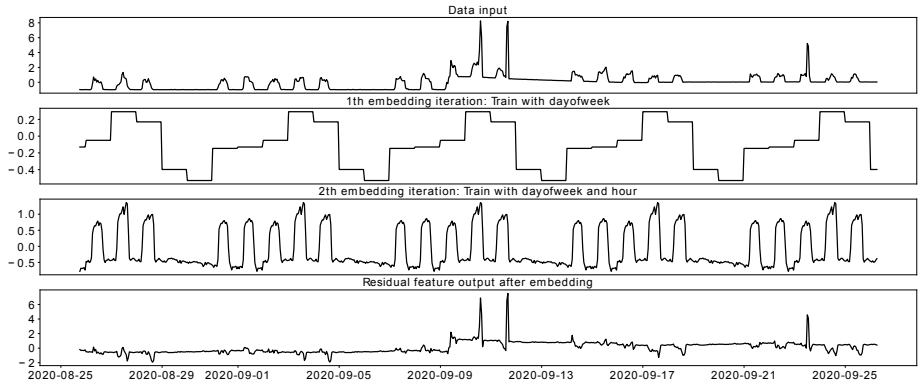}
		\caption{Data set 2}
		\label{fig:roc_1_b}
	\end{subfigure}
	\caption{Embedding comparison over data sets }\label{fig:emb_example}
		\vspace{-10pt}
\end{figure}

Figure \ref{fig:emb_example} corresponds to plots of the outputs of the algorithm. In particular, Figure \ref{fig:emb_example} shows embedding layers' prediction outputs ($model.predict(x)$ in Algorithm \ref{algo:bem}) for two different data sets of server CPU utilization.
In the figures, embedding layers are incrementally trained with data input from two time categorical variables, namely $dayofweek$ (days of a week) and  $hour$ (hours of a day). In the first iteration, the embedding layer learns weekly patterns using $dayofweek$ (i.e., $F_{1}$) as a categorical variable. Then, it uses the next categorical variable $hour$ (i.e., $F_{2}$) to further improve the model by learning hourly patterns. 
The bottom plots shows the final residual (or irregular) time-series feature after embedding $F_{m}$ in Algorithm \ref{algo:bem}.

Figure \ref{fig:emb_example} (a) shows that CPU utilization follows a highly correlated weekly pattern and that the embedding model accurately captures the weekly and hourly regular temporal pattern through 1st and 2nd iterations. Note that the final residual features capture irregular daily pattern in 2020-09-08 (Monday).  In the bottom plot of Figure \ref{fig:emb_example} (b),  three irregular hourly patterns in  2020-09-10, 09-11, and 09-23 are accurately captured in the residual feature and shown as spikes.

\section{Experimental Evaluation}
    

\subsection{Datasets}
\begin{figure}[ht]
\centering
\subfloat[Internal data set]{\includegraphics[scale=0.23]{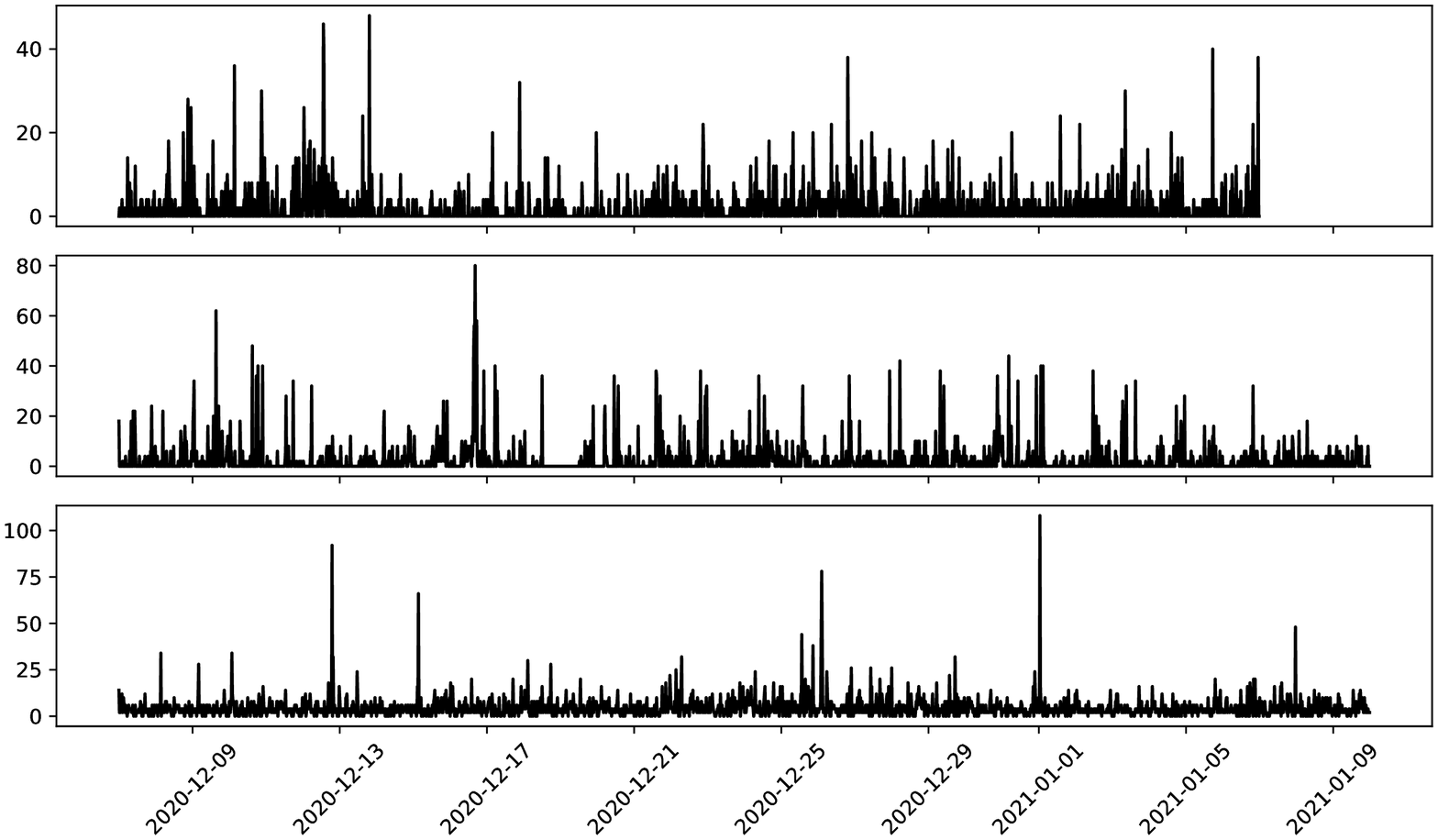}}
\subfloat[Public data set]{\includegraphics[scale=0.23]{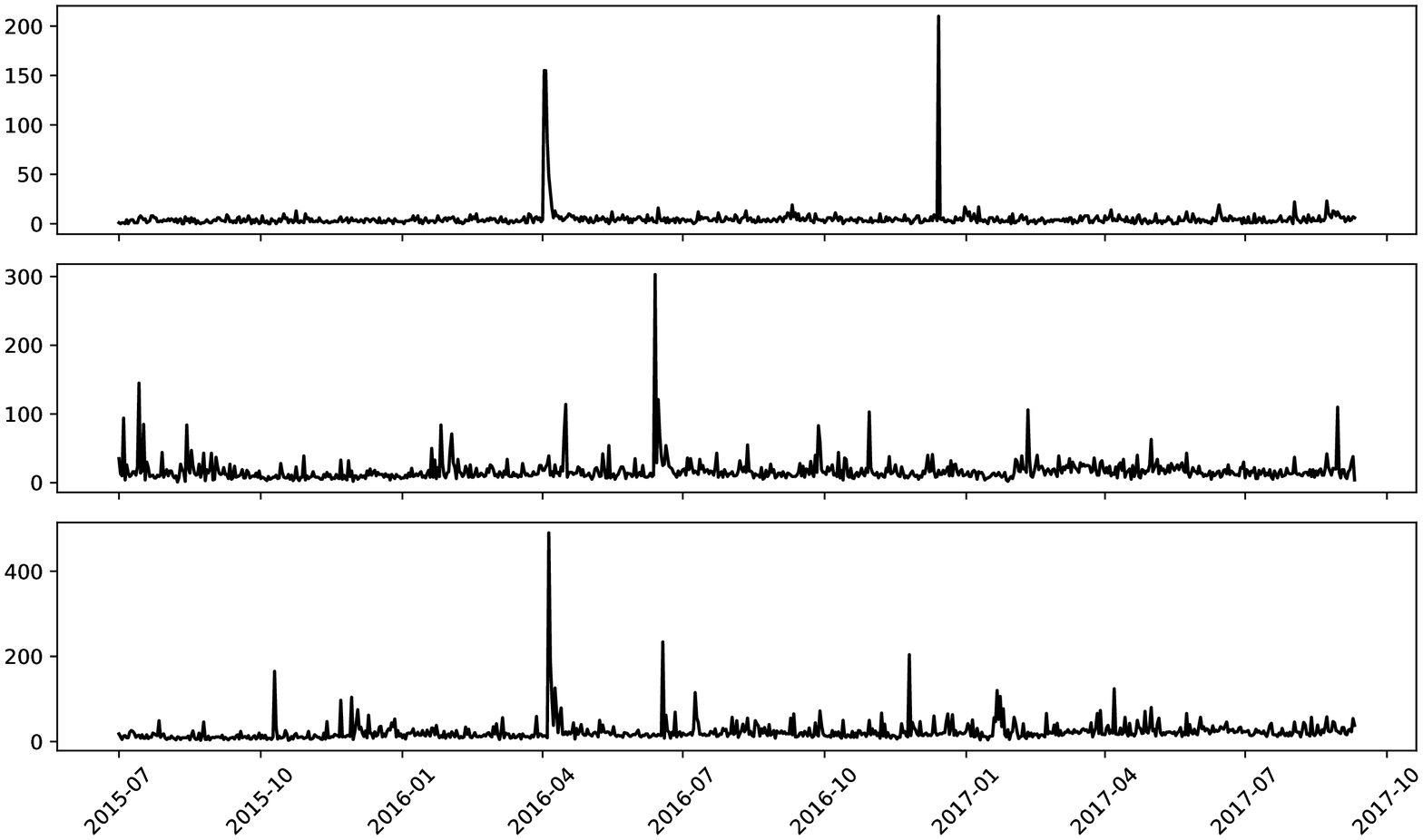}}
\caption{A few sample time-series plots from the internal and public data sets  }
\label{fig:exp_data}
\end{figure}

We conducted experiments on the Wikipedia\cite{wikipedia} data available to the public via the Kaggle platform and internal networking device data. The internal data measures {\em connections per seconds} to a device, a rate-based variable which in general is hard to forecast.  Figure~\ref{fig:exp_data}(a) depicts the time-series for {\em connections per seconds} over a month, from Dec 2020 to Jan 2021. For the sake of simplicity, we use  $I1,\dots,I9$ for representing internal data sets.

The time series in the Wikipedia dataset represents the number of daily views of a specific Wikipedia article during a certain timeframe. Figure~\ref{fig:exp_data}(b) depicts a few time-series plots of the number of accesses of a particular wiki page starting from July 2015 to September 2019. The Wikipedia dataset also represents a rate variable, the number of page access per day. Similar to the internal dataset, we use $P1,\dots,P9$ for representing public data sets. 


\subsection{Model setup} \label{res:pub}
For benchmarking, \texttt{DeepGB} is compared with \texttt{SARIMA} and \texttt{Seq2Seq}. To keep comparisons as fair as possible, the following protocol is used: while employing \texttt{SARIMA}, the implementation in \texttt{statsmodels}~\cite{seabold2010statsmodels} is used. For \texttt{SARIMA}, the best parameters are selected via grid search and \texttt{SARIMA(p=3, d=1, q=0)(P=1, D=0, Q=1)} is the final model employed for our analysis. For \texttt{Seq2Seq}, we started with the architecture in Hwang et al. ~\cite{hwang2019novel} and further tuned to the setting that gives the best empirical performance. The implementation of the \texttt{Seq2Seq} model includes a symmetric encoder-decoder architecture with an LSTM layer of dimension 128, a RepeatVector layer that acts as a bridge between the encoder and decoder modules, a global soft attention layer~\cite{bahdanau2016endtoend} for many-to-one sequence tasks, and a self-attention layer~\cite{lin2017structured}. Also, Time2Vec~\cite{kazemi2019time2vec} and categorical embeddings~\cite{wen2016cat2vec}, that enable the use of categorical features based on the properties of the time-series signature, were added to \texttt{Seq2Seq}. These enhancements improved the performance of \texttt{Seq2Seq}. The proposed \texttt{DeepGB} model consists of two parts - layers in the first half of the model implement embeddings and the later half uses a Boosting model. In particular, \texttt{CatBoost} with $800$ trees of depth $3$ was used as the gradient boosting model for the experiments. The proposed \texttt{DeepGB} model consists of 4 embedding layers followed by a concatenation layer and 4 dense layers with ReLu activation function after each layer. A dropout layer is also added to prevent overfitting. Time2Vec~\cite{kazemi2019time2vec} and categorical embedding~\cite{wen2016cat2vec} were also used in \texttt{DeepGB}. Additionally, Root Mean Square Propagation with step size=0.0002 was used as the optimization algorithm for the experiments. For CatBoost, the implementation on \texttt{scikit-learn 0.24.1} with default parameter setting was used. 


For all the methods, 30 days and 3 days of data were used for training and testing, respectively. Only \texttt{SARIMA} used 14 days of data for training, as its time for convergence increases significantly with the length of data. The widely used metric for time-series error analysis, Symmetric mean absolute percentage error (SMAPE), is employed to evaluate forecast accuracy in the experiments~\cite{armstrong1985long}. SMAPE is the main metric used
in the M3 forecasting competition~\cite{M3}, as it computes the relative error measure that enables comparison between different time series, making it suitable for our setting. 
\subsection{Results}

The results of our experiments for the task of time-series forecasting are summarized in Tables~\ref{tab:res-internal} and \ref{tab:res-public} for the Wikipedia and internal data sets, respectively. 
 Table~\ref{tab:res-internal} shows SMAPE results for \texttt{DeepGB}, Seq2Seq and \texttt{SARIMA}. Bold font indicates the lowest SMAPE for the given dataset. It can be observed that \texttt{DeepGB} is significantly better than \texttt{SARIMA} and Seq2Seq in the majority of the cases. These results help in stating that \texttt{DeepGB} is on par or significantly better than state-of-the-art approaches, in real domains, at scale.  The SMAPE scores presented in Table~\ref{tab:res-public} indicate that \texttt{DeepGB} outperforms the other two methods in 6 out of 8 data sets. A closer inspection of the results suggests that the efficiency of \texttt{DeepGB} is significantly higher (about $3$ times faster) for dataset \textit{P}1.

In general, it can be observed that deep models take more time than statistical methods. However, \texttt{DeepGB} is significantly faster than the other deep model, Seq2Seq, which is reported to have the second-best performance after \texttt{DeepGB}. The results indicate that the proposed \texttt{DeepGB} approach does not sacrifice efficiency (time) for effectiveness (error percentage).
\begin{table}[ht]
\centering
\caption{SMAPE(lower $\Rightarrow$ better) and training time(lower $\Rightarrow$ better) for \texttt{DeepGB}, \texttt{Seq2Seq} and \texttt{SARIMA} methods on the internal dataset.}
\label{tab:res-internal}
\begin{tabular}{|c|c|c|c|c|c|c|}
\hline
\multirow{2}{*}{Dataset} & \multicolumn{3}{c|}{Error (SMAPE)} & \multicolumn{3}{c|}{Training time(secs)} \\ \cline{2-7} 
 & \texttt{DeepGB} & \texttt{Seq2seq} & \texttt{SARIMA} & \texttt{DeepGB} & \texttt{Seq2Seq} & \texttt{SARIMA} \\ \hline
I1 & \textbf{1.04} & 6.12 & 1.44 & 10.38 & 85.80 & 3.19 \\
I2 & 1.75 & 2.00 & \textbf{1.10} & 9.40 & 86.48 & 6.32 \\
I3 & \textbf{5.44} & 21.95 & 20.77 & 8.66 & 86.63 & 5.99 \\
I4 & \textbf{2.71} & 6.92 & 29.59 & 8.28 & 87.63 & 2.49 \\
I5 & 6.33 & \textbf{6.11} & 6.58 & 8.74 & 81.76 & 5.64 \\
I6 & \textbf{4.59} & 9.39 & 11.01 & 8.16 & 82.28 & 3.10 \\
I7 & \textbf{6.98} & 19.03 & 17.61 & 11.72 & 193.54 & 5.00 \\
I8 & 6.81 & \textbf{4.45} & 17.61 & 13.35 & 195.19 & 4.95 \\
I9 & 61.08 & 62.97 & \textbf{59.14} & 12.95 & 196.26 & 3.37 \\ \hline
\end{tabular}
\end{table}


\begin{table}[H]
\centering
\caption{SMAPE(lower $\Rightarrow$ better) and training time(lower $\Rightarrow$ better) for \texttt{DeepGB}, \texttt{Seq2Seq} and \texttt{SARIMA} methods on the public dataset (wikipedia.org\_all-access\_spider).}
\label{tab:res-public}
\begin{tabular}{|c|c|c|c|c|c|c|c|}
\hline
\multirow{2}{*}{Dataset} & \multirow{2}{*}{Original Series} & \multicolumn{3}{c|}{Error (SMAPE)} & \multicolumn{3}{c|}{Training time(secs)} \\ \cline{3-8} 
 &  & \texttt{DeepGB} & \texttt{Seq2Seq} & \texttt{SARIMA} & \texttt{DeepGB} & \texttt{Seq2Seq} & \texttt{SARIMA} \\ \hline
P1 & 2NE1\_zh & 7.98 & \textbf{7.93} & 16.12 & 5.86 & 17.04 & 0.22 \\
P2 & 3C\_zh  & 16.85 & \textbf{6.11} & 15.43 & 6.40 & 20.71 & 0.36 \\
P3 & 4minute\_zh & \textbf{2.34} & 4.38 & 5.90 & 6.22 & 20.36 & 0.36 \\
P4 & 5566\_zh & \textbf{4.39} & 8.01 & 12.44 & 6.05 & 20.73 & 0.30 \\
P5 & AND\_zh & \textbf{5.36} & 13.16 & 25.86 & 8.29 & 24.51 & 0.29 \\
P6 & AKB48\_zh & \textbf{2.27} & 5.81 & 6.78 & 6.30 & 19.48 & 0.34 \\
P7 & ASCII\_zh & \textbf{2.64} & 8.90 & 7.93 & 7.46 & 20.53 & 0.23 \\
P8 & Ahq\_e-Sports\_Club\_zh & \textbf{3.44} & 5.02 & 12.54 & 6.26 & 21.07 & 0.27 \\ \hline
\end{tabular}
\end{table}

\section{Conclusion}
Time-series forecasting is a central problem in machine learning that can be applied to several real-world scenarios, such as financial forecasts and weather forecasts. To the best of our knowledge, this is the first work employing gradient boosting of deep models, infused with embeddings, in the context of time-series forecasting. To validate the performance of the proposed approach, we evaluated \texttt{DeepGB} on a public Wikipedia data set and an internal networking device data set. The experimental results showed that \texttt{DeepGB} outperforms \texttt{SARIMA} and \texttt{Seq2Seq} using SMAPE as performance metric. It was also shown in the empirical evaluations that the proposed method scales well when applied to standard domains as it offers a faster convergence rate when compared to other deep learning techniques. 

Finally, this paper opens up several new directions for further research. Extending the proposed approach to multivariate time-series forecasting and deriving theoretical bounds and convergence properties for \texttt{DeepGB}, remain open problems and are interesting to us from a practical application standpoint.

\bibliographystyle{splncs04}
\bibliography{references}
\end{document}